# Carbon price fluctuation prediction using blockchain information：

## A new hybrid machine learning approach

WANG H.

PANG Y.

SHANG D.* (corresponding author)

**Abstract**

In this study, the novel hybrid machine learning approach is proposed in carbon price fluctuation prediction. Specifically, a research framework integrating DILATED Convolutional Neural Networks (CNN) and Long Short-Term Memory (LSTM) neural network algorithm is proposed. The advantage of the combined framework is that it can make feature extraction more efficient. Then, based on the DILATED CNN-LSTM framework, the L1 and L2 parameter norm penalty as regularization method is adopted to predict. Referring to the characteristics of high correlation between energy indicator price and blockchain information in previous literature, and we primarily includes indicators related to blockchain information through regularization process. Based on the above methods, this paper uses a dataset containing an amount of data to carry out the carbon price prediction. The experimental results show that the DILATED CNN-LSTM framework is superior to the traditional CNN-LSTM architecture. Blockchain information can effectively predict the price. Since parameter norm penalty as regularization, Ridge Regression (RR) as L2 regularization is better than Smoothly Clipped Absolute Deviation Penalty (SCAD) as L1 regularization in price forecasting. Thus, the proposed RR-DILATED CNN-LSTM approach can effectively and accurately predict the fluctuation trend of the carbon price. Therefore, the new forecasting methods and theoretical ecology proposed in this study provide a new basis for trend prediction and evaluating digital assets policy represented by the carbon price for both the academia and practitioners.

# 1. Introduction

Energy transformation for sustainability is essential for human civilization (Yang et al., 2022). Coase Theorem internalizes the economic problem of externalities and provides a new mechanism for the rational allocation of greenhouse gas emissions between economies and enterprises within economies, resulting in the emergence of the carbon emission trading market (Batten et al., 2021). At present, there are several carbon emission trading systems online in the world, among which the European Union Emission Tradings Scheme (EU ETS) is the largest and most mature (Li et al., 2022; Dutta, 2018). At the same time, six years after the signing of the Paris Agreement, China has announced to try to fulfill the goal of carbon peak and carbon neutrality by 2030 and 2060, respectively. It clarifies the stage goals of the "energy revolution" and requires the country and enterprises to make more solid and active efforts for low-carbon energy transformation. China has set up eight pilot carbon markets and officially launched carbon emission trading in the power generation sector in 2021. The most important factor in the carbon trading market is the carbon emission trading price (referred to as carbon price), which directly affects the participation willingness, cost, and future development of enterprises, and then affects the overall emission reduction effect and climate change impact of a country or region.

With the continuous development of the carbon market, carbon price, as one of the core indicators of the carbon financial market, has become increasingly prominent in its financial attributes (Çepni et al., 2022). Predicting carbon prices helps the development of carbon finance markets. The inability to accurately predict carbon prices is a huge obstacle to scientific decision-making by investors and regulators (Estrada et al., 2022; Rodríguez-González et al., 2022). How to investigate the influencing factors of carbon emission trading prices from macro and micro perspectives has become an important topic that many academics, industries, and stakeholders pay attention to.

Carbon trading price forecasting has received high attention in energy and environmental management research and is generally considered one of the most challenging time series forecasting problems (Hua et al., 2022; Kim et al., 2022; Kumari and Toshniwal, 2021; Mota et al., 2022).

Carbon trading price prediction involves significant fluctuations of a large number of unpredictable factors, which can be classified as a nonlinear system, leading to complex time dependence (Batten et al., 2021; Dutta, 2018). The previous researches mainly focus on three aspects: the selection of forecasting paradigm and the influencing factors. In terms of the selection of forecasting paradigm, univariate forecasting refers to the application of analytical techniques to forecast nonlinear time series. Multivariate forecasting means that the carbon price forecasting model not only includes historical carbon price data but also includes external factors affecting the change in the carbon price, which usually include macroeconomic indicators, and energy price indicators (Byun and Cho, 2013; Xu et al., 2020; Wang et al., 2022). Moreover, apart from univariate and multivariate forecasting methods based on statistics and econometrics, machine learn-based research and hybrid forecasting models are analyzed.

Previous studies on carbon trading price prediction and volatility trend analysis provided important references for this study. However, there are still some problems in the prediction accuracy and the selection of multivariate forecasting indicators in carbon price fluctuations. Specifically, first, the studies based on statistical and econometric models have limitations in fitting nonlinear features. Second, in view of system science, the carbon price forecast is inevitably affected by external factors (Kim and Lee, 2021; Wand and Wang, 2021). However, in previous studies, the external influencing factors of the carbon price are not only macroeconomic indicators, but trade, energy prices, and subsidy indicators. At the same time, due to the decentralized trust mechanism, blockchain has been widely used in the fields (Chong et al., 2019; Kim and Lee, 2021). Previous studies have shown that blockchain information indicators significantly correlate with energy prices, such as crude oil prices (Kim and Lee, 2021). However, it is almost not found that the research on the influencing factors of the carbon price, which is also an energy-related indicator, includes the blockchain information indicator. Third, the advancement of machine learning, especially deep learning, and recurrent neural networks represented by Long Short-Term Memory (LSTM) and Convolutional Neural Networks (CNN) show good prediction ability in time series models due to the characteristics of good nonlinear and time lag fitting. However, CNN still have some limitations in the selection of multi-variable predictors and feature extraction. For example, although CNN can be used to extract features, feature extraction with too small fine granularity weakens the effectiveness and fails to extract the feature with optimal value effectively regarding extraction efficiency.

To sum up, in order to overcome the above problems, this paper proposes a novel DILATED CNN-LSTM framework and integrates the Ridge Regression (RR) index screening method (Fan et al.,2009; Matsui and Konishi, 2011; Jung and Park, 2015; Sun et al., 2022) and forecast the carbon price. This RR-DILATED CNN-LSTM hybrid model has the advantages of potential fitting and predicting the fluctuation trend of the carbon price. Specifically, in terms of indicator selection, a core problem in machine learning is to implement algorithms that not only conduct well on training data but also generalize on new inputs. The RR method adopted in this paper has advantages in index selection. Ridge regression is a biased estimation regression method (Fan et al.,2009; Sun et al., 2022). At the cost of partial information loss and precision reduction, a more realistic and reliable L2 regularization method can be obtained. This method minimizes the mean square deviation of the model, thus ensuring the stronger generalization ability of the model. Therefore, the penalty term of RR can screen effective regularization of external influencing factors in macroeconomic indicators, energy indicators, and blockchain information indicators in the small data set in this paper. This paper also verified that the sparsity property derived from L1 regularization SCAD (Smoothly Clipped Absolute Deviation Penalty) has been widely used in the feature selection mechanism as a comparison of L2 regularization. In this paper, RR and SCAD are used to regularize the indicators of external factors that predict carbon price fluctuations. Then DILATED CNN algorithm is applied to extract features more effectively through the expansion and change of the convolution kernel. Finally, LSTM is used to fit the time series data of carbon price fluctuation and carry out the prediction research.

In response the above issues, this study contributes to the literature are as follows: First, this paper proposes a novel DILATED CNN-LSTM framework to predict the trend of carbon price fluctuations. Previous scholars proposed DILATED CNN can extract features more effectively by expanding and changing the convolution kernel to avoid extraction efficiency of CNN. In other words, DILATED CNN can make feature extraction more efficient based on CNN. Second, the combined framework advantage is not limited to CNN extraction of feature vectors. LSTM, as a serial model, can analyze financial time series data. Compared with traditional statistical time series models such as GARCH, LSTM can improve prediction accuracy. These characteristics combine to facilitate the prediction of carbon price prices and volatility trends. Third, based on the comprehensive analysis of previous literature, effective and rich indicators are selected to carry out fitting and prediction, including both internal characteristics of the carbon price and external indicators (such as the S&P 500 index, NASDAQ index, etc.). In particular, when screening indicators, this paper refers to the characteristics of high correlation between energy price and blockchain information in previous literature, and primarily includes indicators related to blockchain technology. Morevoer, in this paper, the dataset containing a large amount of data is used to carry out prediction, regularization methods are used to select indicators, namely Ridge regression (RR) as the reliable L2 regularization method and Smoothly Clipped Absolute Deviation Penalty (SCAD) as L1 regularization regularization method.

Then, we builds a dataset based on carbon prices, macroeconomic indicators, and blockchain information data. Through experiments, by comparing models, the effectiveness and accuracy of prediction could improved. This study provides a new basis for carbon price pricing mechanism and environmental management, carbon price volatility prediction. It also provides a necessary reference for policy making, portfolio management and optimization, and other activities. The remainder of this study is organized as follows. The second part reviews the relevant research literature. The methodology is expected to be explained in Section 3 and the results are discussed in Section 4. Section 5 presents the conclusion, significance, limitations, and future research directions of this study.

## 2. Literature review

Carbon trading price prediction has received high attention in energy and environmental management research. Previous studies on carbon trading price prediction and volatility trends mainly focus on three aspects: prediction paradigm selection, influencing factors, and specific prediction model types, which provide an important reference for this study. Specifically, the first and second category is based on statistics and econometrics, including univariate and multivariate forecasting, and the third category is machine learn-based research and mixed forecasting model research, which also includes univariate and multivariate forecasting. In multivariate forecasting, external influencing factors usually include macroeconomic indicators, energy indicators, and climate indicators denote significance.

Analysis of the research based on statistics and econometrics to build models, first category is the univariate forecasting study. Byun and Cho (2013) tested the three methods of the GARCH-type model, implied volatility model, and K-nearest neighbor model to evaluate the prediction ability of carbon option price volatility. The results show that the GARCH-type model performs better than the implied volatility and K-nearest neighbor models. To detect possible outliers in the European Union Quota (EUA) market, Dutta(2018) applied the GARCH-jump model to the assessment of abnormal values and fluctuations after discovering and removing these extreme points. The results show that the carbon emission price is highly sensitive to the implied volatility of the oil market, and the crude oil volatility index has an asymmetric impact on the EUA market.

Secondly, the research of multivariate forecasting models based on statistics and econometrics are analyzed. To measure the shadow pricing of carbon emission reduction, Wang et al., (2022) proposed a non-parametric method and results show that the gap between the observed carbon shadow price and the best carbon shadow price is significantly different among countries in the world, and the climate change agreement has an impact on global environmental performance. Batten et al., (2021) used multivariate statistical analysis and the results show that the model based on energy prices alone can only explain 12% of the variation in carbon prices, and weather variables do not affect carbon prices. Lin and Xu (2021) used a data-driven non-parametric additive regression model and the results show that the real economy and natural gas price also have a nonlinear effect on the carbon price.

Subsequently, machine learning-based research and hybrid prediction model research are analyzed. It mainly involves traditional machine learning and deep learning. In addition to using artificial intelligence algorithms, hybrid models usually integrate methods from many fields, such as signal processing, applied mathematics, and statistics. On the one hand, the univariate prediction research is analyzed. Li et al., (2022) proposed a carbon price hybrid model by optimizing variational mode decomposition, and results show that a hybrid prediction model has a certain prediction effect. Xu et al. (2020) used an extreme learning machine algorithm and the results show that the prediction model has certain robustness in the face of random samples. On the other hand, machine learning-based multivariable hybrid prediction model research is analyzed. Hao and Tian, 2020 developed a new hybrid framework using an optimal kernel extreme learning machine model and results show that the developed framework outperforms the compared models. Han et al., 2019 proposed a hybrid model of backpropagation neural network and results show that the influence of coal is higher than other factors.

In addition, due to the characteristics of traditional statistical models and econometric models, the nonlinear and time lag characteristics of the fitting and prediction accuracy will be limited. With the development of deep learning, deep learning prediction models and hybrid prediction models have been more applied to time series prediction because of the good nonlinear and time lag characteristics of fitting, and the better ability of fitting time series data than traditional machine learning models. The recurrent neural networks represented by long short-term memory (LSTM) shows good prediction ability in the time series model. For example, in the field of carbon price studied in this paper, Huang et al., (2021) proposed a decomposition integration paradigm with LSTM and results show that the LSTM involved in the model is well suited for predicting the third stage of EU ETS with smaller errors than the single econometric or AI model. Zhou et al. (2019) constructed a hybrid framework with multiple single-step advance predictors to analyze and predict carbon price based on Fully Integrated Empirical Pattern Decomposition with LSTM and the results show that the proposed methods are beneficial to carbon price prediction, but they still need to be optimized for practice.

However, recurrent neural networks represented by LSTM still have some limitations in the selection of multi-variable predictors and feature extraction. The first limitation is the selection of effective predictors. From the perspective of system science theory, carbon price fluctuation is inevitably affected by external factors. When forecasting the carbon price, it is often hoped to find the explanatory variable that can predict the dependent variable very effectively. Therefore, previous scholars have provided a feasible method for variable selection of the model by means of regularization through ridge regression and other methods.

The second limitation is feature extraction. Many previous studies have used CNN algorithm for feature extraction and combined it with LSTM to construct the framework. It has been widely used in traditional computers science fields and financial time series forecasting achieving good results. However, although CNN can be used to extract features, it has some limitations in efficiency. For example, the feature extraction with too small fine granularity wastes computing resources and fails to effectively extract the features with optimal value. Previous scholars proposed DILATED CNN based on this problem and DILATED CNN can extract features more effectively by expanding and changing the convolution kernel. To sum up, this article first through the regularization method in screening effective influence factors, then DILATED CNN algorithm through the expansion of the convolution kernels through the LSTM. Thus, the more effectively pattern of extract the features and fitting carbon price fluctuations of time-series data prediction could be improved.

## 3. Methodology

### 3.1 Parameter norm penalty as regularization

One central problem in machine learning is to implement algorithms balancing performance and generalization. Thus, Many strategies are designed to reduce testing errors, which may come at the expense of increasing training errors (Goodfellow et al., 2016;LeCun et al., 2015). These strategies are collectively known as regularization. The parameter norm penalty is often referred to as the L2 parameter norm penalty for weight decay, which is also known as ridge regression L2 regularization making the weight closer to the origin (Goodfellow et al., 2016; Rusk, 2015). L2 regularization allows the learning algorithm to sense the input x with high variance, so the weight of features with small covariance (relative increase variance) to the output target will shrink (Goodfellow et al., 2016). While L2 weight decay is the most common form of weight decay, we can also use other methods to limit the size of model parameters, one option is to use L1 regularization. In contrast to L2 regularization, L1 regularization will produce more sparse solutions, where sparsity means that some parameters in the optimal value are compressed to 0 (Goodfellow et al., 2016;LeCun et al., 2015). For example, LASSO (Least Absolute Shrinkage and Selection Operator) model combines the L1 penalty with the linear model and uses the Least squares cost function. Sparse properties derived from L1 regularization have been widely used in feature selection mechanisms. Feature selection can simplify machine learning problems by selecting meaningful features from a subset of available features(Goodfellow et al., 2016).

Lasso method overcomes the shortcomings of traditional statistical methods in selecting models and is widely paid attention to(Matsui and Konishi, 2011; Jung and Park, 2015). Subsequently, Lasso and its various optimization methods have been widely used in the research of variable

selection(Matsui and Konishi, 2011). Although LASSO has many good properties, there are still some drawbacks. If you have a bunch of variables in the data that have strong pairwise interactions, you tend to pick one out of the bunch and not care which one is picked. Among many improved methods, the method of Smoothly Clipped Absolute Deviation Penalty is essential. Compared with Lasso, the method of SCAD reduces the Deviation of parameter estimation (Matsui and Konishi, 2011; Jung and Park, 2015). Therefore, SCAD has been widely used since it was proposed. Because S CAD has the advantage of selecting corresponding variables, this method is used in many previous kinds of research. In this paper, Ridge Regression and SCAD methods are also used to screen the external influencing factors of price prediction and conduct L2 and L1 regularization.

### 3.1.1 Ridge Regression

Multiple linear regression model is a basic model which has a good ability to fit general statistical problems. At the same time, due to the existence of closed form solutions, the problem can be solved very quickly. Multiple linear regression model is used to express the relationship between target data y and observed data x1,x2, xm, namely:

$$y = \beta_0 + \beta_1 x_1 + \cdots + \beta_m x_m + \varepsilon \quad (1)$$

Where, B0,B1 and Bm are regression coefficients and ε is random error. Assuming n groups of target data and observation data, equation (2) below can be established to solve the regression coefficients.

$$\begin{cases} y_1 = \beta_0 + \beta_1 x_{11} + \cdots \beta_m x_{1m} + \varepsilon_1 \\ y_2 = \beta_0 + \beta_1 x_{21} + \cdots \beta_m x_{2m} + \varepsilon_2 \\ \quad \vdots \\ y_n = \beta_0 + \beta_1 x_{n1} + \cdots + \beta_m x_{nm} + \varepsilon_n \end{cases} \quad (2)$$

Expressed in matrix form:

$$Y = X_p + E \quad (3)$$

Where, Y=(y,y,y,)",p=(B,B,p,)",s=(g,s-,)". According to the Gauss-Markov theorem, the best linear unbiased estimation of regression coefficient is the least square estimation in the linear regression model with error satisfying zero mean, and correlation. Therefore, the regression model of equation (3) above can be completely written as follows:

$$y = X\beta + \varepsilon, E(\varepsilon) = 0, V_{ar(\varepsilon)} = \sigma^2 \quad (4)$$

Using the least square method to obtain the regression coefficient:

$$p = (x''x)'' x'y \quad (5)$$

Substitute Equation (3) into Equation (5), and the regression coefficient becomes:

$$p = (x'x)'' x'y$$
$$= (x'x)' x''xp + (x'x)' x'e \quad (6)$$
$$= p + (x'x)' x'e$$

$$Y = X\beta + \varepsilon \quad (7)$$

The mean square deviation (MSE) was used to evaluate the robustness of the prediction ability of the model. The smaller the mean square deviation, the larger the prediction accuracy. Assuming $X^T X$ invertible, $\lambda_1, \lambda_2, \cdots \lambda_p$ is $X^T X$, then $(X^T X)^{-1}$ of the characteristic root of is $\lambda_1^{-1}, \lambda_2^{-1}, \cdots \lambda_p^{-1}$,

$$MSE(\hat{\beta}) = \sigma^2 \sum_{i=1}^{p} \lambda_f^{-1} \quad (8)$$

If X'X has a relatively small characteristic root, MSE(P) will be large. In this way, the parameters obtained by unbiased estimation are not a good estimate. Thus, Ridge Regression was applied as more realistic and reliable regularization with regression coefficients obtained at the expense of partial information loss and precision reduction (Bertinetto et al., 2018). The ridge regression model has the smallest mean square deviation, thus ensuring a stronger generalization ability of the model. It is widely used and continuously developed not only in basic machine learning but also in the field of index screening (Bertinetto et al., 2018).

In terms of the principle of ridge regression, the matrix representation also refers to Equation (3). The objective of ridge regression is to find a function f(x)=p'x, and the existence of β minimizes the residual function value between the objective function and the original response value. The residual function is expressed as follows:

$$\min_{\beta} (Y - f(x)) = \min_{\beta} \sum_{i=1}^{m} (y_i - \beta^T x_i)^2 \quad (9)$$

In order to minimize the regression model MSE, the generalization ability of the model is stronger. Ridge regression proposes to introduce a to limit the sum of all p, and by adding a penalty term, β can be guaranteed to fluctuate in a certain range, that is, the residual function becomes:

$$\min_{\beta}\left(\sum_{i=1}^{m}(y_i - \beta^T x_i) + \lambda \|\beta\|^2\right) \quad (10)$$

The regression coefficient obtained by ridge regression method is as follows:

$$\hat{\beta} = \arg\min_{\beta}\left(\sum_{i=1}^{m}(y_i - \beta^T x_i)^2 + \lambda \|\beta\|^2\right) \quad (11)$$

Parameter values of ridge regression can be obtained as follows:

$$\beta = (X^T X + \lambda I)^{-1} X^T Y \quad (12)$$

Where, I is the identity matrix, and in is the adjustment parameter controlling the penalty intensity. Ridge regression is to solve the optimization model shown in Equation (11). The loss function of this model is smooth and derivable everywhere, and the closed-form solution shown in Equation (12) can be obtained through the necessity condition for the existence of extreme values.

### 3.1.2 Smoothly Clipped Absolute Deviation Penalty

LASSO estimation can be understood as a penalty that minimizes the sum of squared residuals by adding a coefficient. It is the natural nature of having a penalty similar to the soft threshold principle that makes it possible for both continuous compression and automatic selection of variables. Although LASSO has been widely concerned and applied since it was proposed, it still has some inherent defects. For example, one of the biggest problems is the shortcoming that Lasso can only select n variables at most when dealing with data construction whose dimension is larger than the sample size (p > n). And LASSO doesn't have the same performance as oracle procedures. SCAD method improves the above problems. By appropriately choosing regularization parameters, the SCAA method is shown to perform the proposed estimator as well as the oracle procedure in terms of variable selection (Matsui and Konishi, 2011; Jung and Park, 2015).Let the coefficient of the model be represented by β, and its corresponding loss function $l(\beta)$, where β is a D-dimensional column vector, then the penalized likelihood function with respect to the parameters is:

$$l(\beta) + \sum_{i=1}^{d} p_{\lambda_i}(|\beta_i|) \quad (13)$$

When $l(\beta) = (y - X\beta)^2$, $p_{\lambda_i}(|\beta_i|) = \lambda |\beta_i|^q$, known as L1 regularization.

For the linear model, the model selection can be expressed as the following problem:

$$\hat{\beta} = \arg\min_{\beta} \|y - X\beta\|^2 + \lambda \sum_{i=1}^{d} |\beta_i| \quad (14)$$

There are two processes involved: finding significant variables and estimating the corresponding coefficients. When using Lasso and its related methods, these two processes go hand in hand. Lasso actually amounts to considering the following questions:

$$\hat{\beta} = \arg\min \hat{\beta} = \arg\min_{\beta} \|y - X\beta\|^2 \quad \text{when} \sum_{i=1}^{d} |\beta_i| \leq t \quad (15)$$

This approach makes the later inequalities effectively restrict the parameter space. For the sake of description, it is assumed below that the data should be standardized in advance.

(1) Lasso is estimated as:

$$\hat{\beta} = \arg\min_{\beta} \|y - X\beta\|^2 + \lambda \sum_{i=1}^{d} |\beta_i| \quad (16)$$

(2) SCAD is estimated as:

$$\hat{\beta}^{SCAD} = \arg\min_{\beta} \|y - X\beta\|^2 + \sum_{i=1}^{d} p_{\lambda}(|\beta_i|) \quad (17)$$

SCAD (Smoothly Clipped Absolute Deviation Penalty), the form of the regression penalty term can be expressed as:

$$p_{\lambda}(|\beta_j|) = \begin{cases} \lambda |\beta_j| & 0 \leq |\beta_j| < \lambda; \\ -\left(|\beta_j|^2 - 2a\lambda |\beta_j| + \lambda^2 \right) / [2(a-1)], & \lambda \leq |\beta_j| < a\lambda; \\ (a+1)\lambda^2 / 2 & |\beta_j| \geq a\lambda; \end{cases} \quad (18)$$

Pair of penalty functions $p_{\lambda}(|\beta|)$ come to $\beta$

$$p'_\lambda(|\beta_j|) = \lambda\left\{I(\beta_j \leq \lambda) + \frac{(a\lambda - \beta_j)}{(a-1)\lambda} + I(\beta_j > \lambda)\right\} \quad (19)$$

When $a > 2, \beta_j > 0$, estimate $\hat{\beta}_j^{SCAD}$

$$\hat{\beta}_j^{SCAD} = \begin{cases} sign(\hat{\beta}_j^0)(|\hat{\beta}_j^0 - \lambda|), & |\hat{\beta}_j^0| \leq 2\lambda; \\ \left[(a-1)\hat{\beta}_j^0 - sign(\hat{\beta}_j^0)a\lambda\right]/(a-2), & 2\lambda \leq |\hat{\beta}_j^0| < a\lambda; \\ \hat{\beta}_j^0, & |\hat{\beta}_j^0| \geq a\lambda. \end{cases} \quad (20)$$

Therefore, when Lasso selects variables in the face of relatively small coefficients, it will impose relatively large compression, which will lead to a certain degree of overcompression. The penalty term of SCAD is a piecewise function, and the obtained SCAD estimation is also in piecewise form, which solves this problem to some extent. Compared with Lasso, SCAD reduces the deviation of parameter estimation.

**3.2 DILATED CNN and LSTM**

The background of the generation of CNN is based on biological research on the generation and processing mechanism of visual stimuli. In the earliest computer science field, CNN network structure design is usually composed of convolutional layer, pooling layer used for downsampling and fully connected layer (Bengio et al., 2019). The CNN network first extracts the features of the input data through the convolutional layer, then extracts the extracted features through the pooling layer, and finally outputs the extracted features after adding a fully connected layer. CNN belongs to a kind of forward neural network. CNN network is characterized by the introduction of convolutional layer structure, convolutional layer with the help of a filter-like convolution kernel mechanism to extract features in the input matrix. Convolution kernels in general will be far less than the dimension of the input features, so each neuron in the convolution layer will catch up with just a part of the network layer neurons interact, we call this way for the local connection or sparse interaction is the benefits of this treatment will greatly reduce the complexity of the model and the calculation time, and can also be improved to some extent fitting phenomenon. However, this design has some limitations, that is, the pooling operation and the use of down-sampling will lose a certain amount of information, and the receptive field will also be limited.

DILATED CNN first appeared in the field of computer vision, which was born in order to overcome the pooling operation and the loss of a certain amount of information by down-sampling (Yu and Koltun, 2015). In the field of image segmentation computing science, images are input into CNN, and convolution and pooling of images are performed, which can reduce the image size and increase the receptive field at the same time (Muhammad et al.,2021). However, as image segmentation prediction has a technical characteristic, it is the output of pixel-wise. The pooling operation enables each pixel prediction to see the larger receptive field information. In this way, the pooling smaller image sizes are upsampling to the original image sizes for prediction (equivalent to deconvolution operation). So in the process of first decreasing and then increasing the size, some information must be lost. With DILATED CNN design, more information can be obtained with larger receptive field without pooling link.

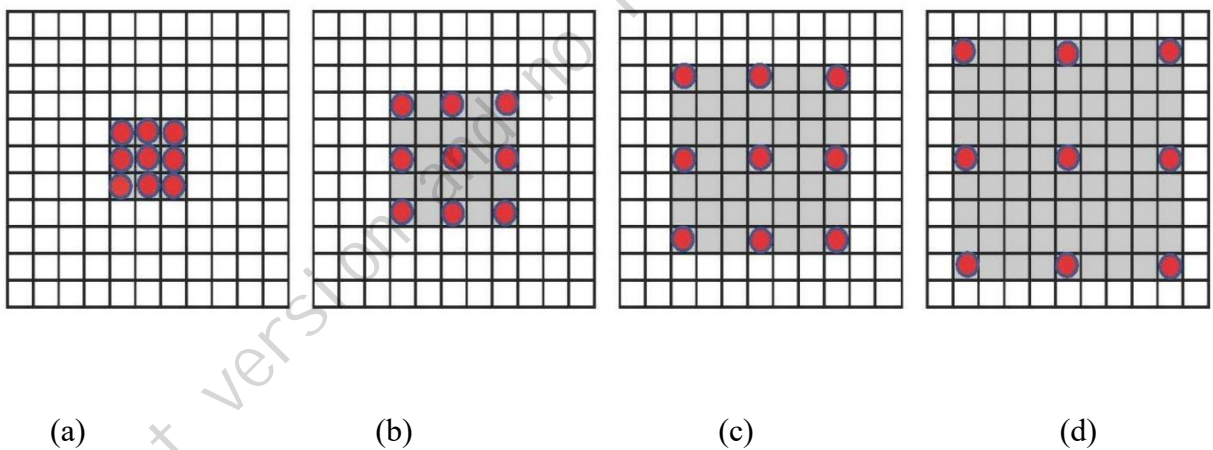

(a)          (b)          (c)          (d)

Figure 1

(a) Figure corresponds to a convolution kernel of 3x3, which is DILATED convolution with DILATED rate of 1. (b) Although the actual convolution kernel size is still 3x3, when the DILATED rate is 2, that is, for a 5*5 image patch, only 9 red points are convolved with the 3x3 kernel, and the remaining points are omitted. So far, the size of kernel is equivalent to 5*5, but only the weight of 9 points in the figure is not 0, and the rest are 0. Similarly, figures (c) and (d) correspond to convolution kernels of 3x3, which are DILATED convolution with DILATED rates of 3 and 4. In other words, for a 7*7 and 9*9 image patch, only 9 red points are convolved with 3x3 kernel, and the remaining points are skipped. So far, the size of kernel is equivalent to 7*7 and 9*9, but only the weight of 9 points is not 0. The receptive field of traditional convolution is linearly related to the number of layers. Compared with traditional convolution, the receptive field of DILATED convolution expands and grows rapidly. The advantages of DILATED convolution can be summarized as follows: firstly, the internal data structure is retained without pooling operation; secondly, without pooling operation loss information, down-sampling is avoided and the receptive field is enlarged. Let each convolution output contain a wide range of information. Therefore, the algorithm has been paid much attention and applied in natural language processing field, semantic understanding and segmentation. But the application in time series forecasting is still very limited.

Meanwhile, in order to effectively explore the dependence relationship between various parts of sequence data, RNN connects neurons in series, so that each neuron has a certain memory and can store the information of the previous input sequence. In this way, sequence data is continuously compressed so that it can be expressed abstracted (Sun et al., 2022). The structure of LSTM neural network is derived from RNN. LSTM included three gate infrastructure, namely forgetting, input and output gate. And a memory cell to selectively obtain incoming data (Muhammad et al.,2021; Bengio et al.,2019).

In LSTM, the forgetting gate is applied for receiving the data from the previous moment ct−1 to the current time ct point calculating the information retention degree ft according to the output $h_{t-1}$ at the previous moment and the input xt at the current moment. After the sigmoid activation function, the value range of ft is between 0 and 1. The function of the input gate is to selected information to the memory cell ct−1 at the previous time and updated to the new memory cell ct.

$$f_t = sigmoid\left(W_f\left[h_{t-1}, x_t\right] + b_f\right) \quad (21)$$

$$i_t = sigmoid(W_i[h_{t-1}, x_t] + b_i) \quad (22)$$

Where, Wi represents the parameter matrix of the hidden layer output $h_{t-1}$ and the input xt, and bi represents the bias term. Then calculate Zt:

$$z_t = \tanh\left(W_g\left[h_{t-1}, x_t\right] + b_g\right) \quad (23)$$

The updated ct of the current memory cell is:

$$c_t = f_t \cdot c_{t-1} + i_t \cdot z_t \quad (24)$$

The function of the output gate is to calculate current information being output based on the output $h_{t-1}$ and the input $x_t$ :

$$o_t = sigmoid\left(W_o\left[h_{t-1}, x_t\right] + b_o\right) \quad (25)$$

The forward propagation of the circulation body in the LSTM neural network is over, and then the error back propagation is calculated, and the weight coefficients are corrected. The partial derivatives of the parameters passing through each gate element are still calculated, and the weights are more weighted in the direction of gradient convergence (Ribeiro et al., 2021).

### 3.4 Hybrid framework for training and prediction process

The SCAD-DILATED CNN-LSTM hybrid model is shown in Fig.2:

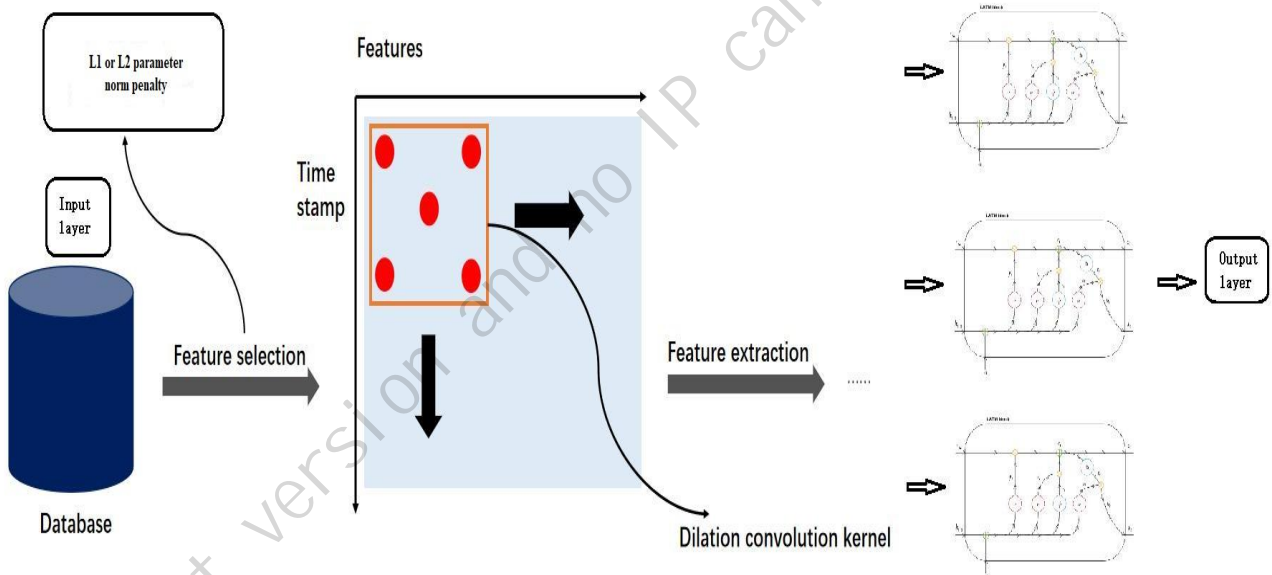

Figure 2

The process of model training consists of the following steps. The first is to normalize the data. Second, in the case of variable selection without group variables, RR and SCAD was used for statistical inference and estimation of the index data, and the variable selection test results were compared. Third, the external factors affecting carbon price and carbon price will be formed into a formal data set after index screening for training and prediction of deep learning models. Fourth, DILATED CNN algorithm is applied to extract features more effectively by expanding and changing the convolution kernel. Fifth, the time series data of carbon price fluctuation is fitted by LSTM to carry out prediction research. After building the network structure of the sixth initial model, parameters including learning rate, batch_size, and Epochs should be set before training the model. The data set is then usually divided into training and testing sets. The seventh calculates the corresponding Error and back Propagation. Update the weight and deviation of each layer back to Step 4 to train if necessary. Finally, the calculation includes input data and standardization, and then the information is input into the proposed network structure DILATED CNN-LSTM after training to obtain the correct output value. To complete the prediction process, the output is restored to the result before standardization.

## 4 Experiments

### 4.1 Data selection and evaluation method

In this study, the predicted carbon price data are selected from Guangzhou Carbon Exchange, China. The Guangzhou carbon trading system is one of the earliest and largest pilot carbon emission exchanges in China. Therefore, it is representative to select the data of the Guangzhou Carbon Exchange among the eight pilot carbon emission exchanges established in China. The data information from April 28, 2017, to August 31, 2021, were selected from the websites of Guangzhou Carbon Exchanges in China. The carbon Price forecast includes the closing price in forecasting time series data. Exogenous variables come from, 53 series, divided into three clusters: macroeconomic series, financial and energy series, and blockchain information series. All indicators are denoted in the appendix table in details.

Specifically, the macroeconomic series includes a total of 13 indicators. For example, SPX Index (USD) and INDU index (USD),. The second aspect is the Global Currency Ratio, such as US Dollar (USD)/China Yuan Renminbi (CNY). The above macroeconomic indicators are referred to previous studies(Kim and Lee, 2021; Jabeur et al., 2021). The financial and energy indicator series comprises 25 indicators. Metals and agricultural communities series includes New York Gold and New York Silver etc. Energy indicators includes WTI New York crude oil, Brent crude oil, China Thermal coal Price Index (point)-Bohai Rim thermal coal price etc. The above financial and energy indicators are derived from previous studies (Lin and Xu, 2021; Kim and Lee, 2021; Wang et al., 2020).

Blockchain information including digital encryption currency trading, on one hand, data from a widely accepted Bittrex exchange, such as Bitcoin and Ethereum session data as the experimental data, each group of data contains six: the opening price, the highest, and the lowest price, closing price, trading volume, market capitalization, a total of 12 indicators. Price data are in U.S. dollars. Blockchain information on the other hand includes energy consumption indicators, and data source blockchain information websites. It specifically includes three indicators, such as the Bitcoin energy consumption data estimator and the minimum amount, and the Ethereum energy consumption data estimator. The above blockchain information indicators are derived from previous studies (Kim and Lee, 2021; Livieris et al., 2021; Jabeur et al., 2021). All external influencing factor variables, namely macroeconomic series, financial and energy series, and blockchain information series indicators are aligned with the carbon price in the data time to construct the dataset.

RR regularization results show that P values of all indicators are less than 0.05, that is, all indicators have significance (see appendix table). Therefore, experimental dataset 1 was generated for the experimental range included by all indicators. Then, dataset 1 is divided into two subsets. 90% is the training set and the remaining 10% the test set. To assess the effectiveness, the proposed method is planned to be compared with CNN-LSTM and DILATED CNN-LSTM under the same training and test set data.

This study introduces a step ahead prediction. All methods are in Python. All experiments are calculated with inteli7-4700h2.6ghz, 256GBs memory and windows10. In order to evaluate the prediction effect of RR-DILATED CNN-LSTM, MSE, MAE and MAPE were used as evaluation criteria. Apart form MSE, Mean Absolute Error (MAE) is a test used to compare predicted values with actual data. This measure is the average of absolute errors. Mean Absolute Percentage Error (MAPE) is the mean absolute percentage error. It is used to evaluate the same set of data from different models. MAPE and MAE are calculated as follows:

$$MAE = \frac{1}{n}\sum_{i=1}^{N}|y_i - \hat{y}_i| \quad (27)$$

$$MAPE = \frac{1}{N}\sum_{i=1}^{N}\frac{|y_i - \hat{y}_i|}{y_i} \quad (28)$$

where N is the number of data in the out-of-sample subset, yi and y estimate, are the real and prediction value, respectively.

The parameter settings are revealed in Table 1. The loss functions are MSE, MAE, MAPE, the optimizer selects Adam, the batch size is 64, the time step is 5, and the learning rate is 0.001.

Table 1 Parameter Settings

| Parameters | Value |
| --- | --- |
| DILATED CNN layer in_channels | 1 |
| DILATED CNN layer layer out_channels | 16 |
| DILATED CNN layer layer kernel size | 3 |
| DILATED CNN layer layer stride | 1 |
| DILATED CNN layer layer padding | 1 |
| Number of hidden units in LSTM layer | 32 |

As an L2 parameter norm penalty, experimental dataset 1 will verify the following models 1-4, RR-CNN, RR-LSTM, RR-CNN-LSTM, and RR-DILATED_CNN-LSTM. Similarly, some indicators are compressed to 0 by the SCAD algorithm, that is, only some indicators have significance. For specific indicators, please refer to Appendix Table. Dataset 2 screened by SCAD is a subset of experimental dataset 1. In other words, experimental dataset 1 (generated by the RR algorithm) is the complete set of experiments in this study, with a total of 53 indicators, and each indicator has more than 1000 time-point data. Experimental dataset 2 (generated by the SCAD algorithm) is a subset of this study, with a total of 23 indicators. As the L1 parameter Norm penalty, experimental dataset 2 will verify model 5, SCAD-DILATED_CNN-LSTM. The implementation process of Model 5 is the same as Model 1-4 except that different experimental datasets 2 are used.

4.2 Analysis of experimental results

The results of the methods proposed with comparison are revealed in Table 2. From Table 2, for Strong correlation featuress, the MSE of LSTM is the largest, and RR-DILATED_CNN-LSTM is the smallest. The performance with L2 parameter norm penalty methods from high to low is RR-DILATED_CNN-LSTM, RR-CNN-LSTM, RR-CNN, and RR-LSTM, considering robust correlation features. For MAE and MAPE, the RR-LSTM is also not satisfactory compared to results of other methods. Similarly, compared with RR-CNN-LSTM, RR-CNN, and RR-LSTM, the MAE and MAPE of RR-DILATED_CNN-LSTM are 0.174 and 0.270 which are the best indicator results than other competing models. Meanwhile, using SCAD as the L1 parameter norm penalty, the results are better than some algorithms, but not better than the RR under the same network results as the L2 parameter norm penalty. Namely, the MAPE of SCAD-DILATED_CNN-LSTM is 0.393 are better than RR-CNN model and RR-LSTM model. However, MSE, MAE and MAPE of SCAD-DILATED_CNN-LSTM model are all higher than RR-DILATED_CNN-LSTM model denoting that RR-DILATED_CNN-LSTM is the best approach in this study. As the financial time series of digital assets and bulk commodities can be seen as the nonlinear system, the carbon prices are inevitably affected by other external factors. The results denote that our proposed method, RR-DILATED_CNN-LSTM, show the most satisfied prediction effect.

Table 2 Experimental Results

| Regularization of parameter norm penalty | | MSE | MAE | MAPE |
|---|---|---|---|---|
| L2 | RR-CNN | 0.058 | 0.227 | 0.399 |
| L2 | RR-LSTM | 0.251 | 0.493 | 1.564 |
| L2 | RR-CNN-LSTM | 0.040 | 0.200 | 0.279 |
| L2 | RR-DILATED_CNN-LSTM | **0.037** | **0.174** | **0.270** |
| L1 | SCAD-DILATED_CNN-LSTM | 0.065 | 0.239 | 0.393 |

## 5.Discussion

This study contributes to the literature in the following ways. First, we find an essential association between blockchain indicators and carbon prices and suggest that blockchain indicators can be used to predict future carbon prices. Previous studies have found that the technical characteristics of distributed computing and storage and blockchain information such as bitcoin and Ethereum prices are significant influencing factors for predicting oil and energy prices (Chong et al., 2019; Kim and Lee, 2021). However, carbon price forecasting studies do not take this view into account. Our study reveal that blockchain information, such as cryptocurrencies price and amount information, Bitcoin energy consumption and the Ethereum energy consumption estimators, are relevant for predicting carbon prices. Given the rise and future value of cryptocurrencies and other blockchain applications, it is expected that blockchain information such as cryptocurrency prices will be important. The results of this study not only help to enhance the accuracy of carbon price prediction but also offer evidence of the advantages of using blockchain information to carbon price time-series prediction.

Secondly, in addition to the blockchain information variables related to carbon price prediction, we also found other variables related to price prediction through regularization, namely RR method application, macroeconomic variables, and energy variables. From the perspective of system science theory, carbon price fluctuation is inevitably affected by external factors. Therefore, when forecasting the carbon price, it is a dilemma to effectively screen the explanatory variables of the predicted dependent variable. In previous studies on influencing factors of the carbon price, few statistical dimensionality reduction methods such as Ridge Regression were used to scientifically screen variables. In this paper, ridge regression is applied to minimize the mean square deviation of the model, thus ensuring a stronger generalization ability of the model (Bertinetto et al., 2018). This paper also uses this method to screen the external factors of price prediction. The results show that the blockchain information indicators, macroeconomic indicators, and energy indicators are included in our research results, which can effectively select variables and maximize the retention of effective predictive indicators, ensuring that the model training can contain as much information as possible. At the same time, the results show that RR acts as L2 regularization prior to SCAD acting as L1 regularization in prediction. Therefore, this paper not only ensures the forecasting efficiency by using RR as L2 regularization, but also provides supporting empirical evidence for regularization in the selection of carbon price forecasting indicators, and extends the application scope of RR theory.

Thirdly, we find that DILATED CNN has a better effect on the feature extraction of the carbon price. In previous studies, especially those involving financial time series, CNN is often used to extract feature vectors. In this way, CNN is involved in pooling smaller information and then upsampling it to the original information vectors for prediction. In the process of decreasing and then increasing the size, some information would be lost. With DILATED CNN design, more information can be obtained with a larger receptive field without pooling links. The advantage of DILATED CNN adopted (Yu and Koltun, 2015; Muhammad et al.,2021) is that on the one hand, the internal data structure is retained without pooling operation and information loss is reduced. Also, down-sampling is avoided, which increases the receptive field and makes each convolution output contain a large range of information. The results reveal that DILATED CNN can extract features more effectively, which is good extension for the construction of the CNN-LSTM framework. More importantly, DILATED CNNS used to be used in the natural language processing field. This study promotes the generalization ability of the DILATED CNN theoretical model and provides empirical evidence for other similar studies.

To sum up, this study proposes a RR-DILATED CNN-LSTM framework as a machine learning approach to predict the trend of carbon price fluctuations. In the past, the CNN-RNN framework is often used to fit sequence models and carry out prediction research. Although the CNN-RNN framework has been applied in many research fields, showing the practical price, the prediction accuracy still needs to be further improved. In this paper, a novel DILATED CNN-LSTM framework is constructed to predict carbon price and trend based on the research framework of DILATED CNN and recurrent neural networks, and the LSTM neural network algorithm is introduced based on the temporal model of the traditional recurrent neural network. The combined framework advantage is not limited to feature vector extraction with CNN, but DILATED CNN can make feature extraction more efficient based on CNN. LSTM, as a serial model prediction, can analyze time series data and improve prediction accuracy compared with traditional statistical time series models such as GARCH (Byun and Cho, 2013; Goodfellow et al, 2016). The results show that the combination of these characteristics facilitates the prediction of carbon price fluctuations. RR-DILATED CNN-LSTM has the highest accuracy compared with the benchmark model.

Thus, the contribution of the RR-DILATED CNN-LSTM framework proposed in this paper can not only provide a basis for the theoretical research, practitioners of the carbon price but also be applied to the practitioners of the carbon price and other energy financial series analysis. For example, one the one hand policymakers can adjust and formulate short-term policies according to the price fluctuation trend prediction. On the other hand, quantitative finance practitioners can better conduct risk management and portfolio management formulation through the association between energy indicators such as carbon price and digital assets based on blockchain. In summary, the novel framework expands the application of the previous deep learning framework. The framework and results also aims to help practitioners forecasting carbon prices in the future which can easily replicate the methods and assessment process.

## 6. Conclusion

This study proposes a novel DILATED CNN-LSTM framework and integrates the RR Regularization method to predict carbon price. This RR-DILATED CNN-LSTM hybrid model has the advantage of fitting and predicting the fluctuation trend of the carbon price. Specifically, this study is among the first to proposes a multi-stage new method based on the relevant theories of machine learning, that is, based on the research framework of DILATED CNN and recurrent neural network, and introduces the LSTM neural network algorithm based on the temporal model of traditional recurrent neural network. The novel DILATED CNN-LSTM framework is constructed to predict carbon prices and trends. The advantage of the combined framework is that DILATED CNN can make feature extraction more efficient based on CNN. LSTM, as a serial model prediction, can analyze financial time series data. Compared with traditional statistical time series models such as GARCH, DILATED CNN-LSTM can improve prediction accuracy. These characteristics combine to facilitate the prediction of carbon price prices and volatility trends. Second, based on the comprehensive analysis of previous literature, effective and rich indicators are selected to carry out fitting and prediction, including both internal characteristics of the carbon price and macroeconomic indicators. In particular, when screening indicators, this paper refers to the characteristics of high correlation between energy price and blockchain information in previous literature, and primarily includes indicators related to blockchain information. Thirdly, this paper adopts a dataset containing a large amount of data to carry out the prediction and adopts the RR method for regularization. The results show that the RR method minimizes the mean square deviation of the model, thus ensuring the stronger generalization ability of the model. In this small sample data set, RR as L2 regularization is better than SCAD as L1 regularization in price forecasting. Therefore, this paper not only provides supporting empirical evidence by using RR as a carbon price predictor for L2 regularization, but also extends the application scope of RR theory. Therefore, this paper retains as much predictive information as possible to construct a model that is realistic and easy to interpret. To sum up, on the one hand, the contribution of the RR-DILATED CNN-LSTM framework proposed in this paper can not only provide a basis for the theoretical research and practitioners of carbon price but also be applied to the academia and practice. On the other hand, the novel framework proposed in this study expands the theoretical perspective of previous machine learning frameworks and improves the generalization ability to exist theoretical models.

This study also has some limitations. There is still improvement for further optimization in the construction of the forecasting framework and indicator screening. For example, although DILATED CNN has advantages in feature selection and information preservation, there is still advancement for optimization. The DILATED CNN kernel is not continuous, that is, not all pixels will be counted. The detection effect of large dilated rate convolution for small volume is not ideal and many information vectors cannot be considered. Even if multi-scale transformation can be done, it will increase the complexity of the network. Thus, one attempt could be Hybrid DILATED Convolution. The dilated rates are changed into different forms, that is, the dilated between different layers is constantly changing. The goal is for the final receiving field to cover the entire area. At the same time, regularization can also be improved such as adding random noise. Therefore, the future research directions are to further improve the prediction accuracy and robustness by building a combination of a more advantageous forecasting framework and regularization methods.